# Evolutionary Computation in Astronomy and Astrophysics: A Review

José A. García Gutiérrez, Carlos Cotta, and Antonio J. Fernández-Leiva

*Dept. Lenguajes y Ciencias de la Computación, ETSI Informática,*
*Campus de Teatinos, Universidad de Málaga,*
*29071 Málaga – Spain*
*{ccottap, afdez} @lcc.uma.es*

In general Evolutionary Computation (EC) includes a number of optimization methods inspired by biological mechanisms of evolution. The methods catalogued in this area use the Darwinian principles of life evolution to produce algorithms that returns high quality solutions to hard-to-solve optimization problems. The main strength of EC is that they provide good solutions even if the computational resources (e.g., running time) are limited. Astronomy and Astrophysics fields often require optimizing problems of high complexity or analyzing a huge amount of data and the so-called complete optimization methods are inherently limited by the size of the problem/data. Reliable analysis of large blocks of data is central to modern astronomical sciences in general. EC techniques perform well where other optimization methods are inherently limited (as complete methods applied to NP-hard problems). For this reason in the last ten years, numerous proposals have come up that apply with greater or lesser success methodologies of evolutional computation to common engineering problems. Some of these problems, such as the estimation of non-lineal parameters, the development of automatic learning techniques, the implementation of control systems, or the resolution of multi-objective optimization problems, have a special repercussion in the fields and EC emerges as a feasible alternative for traditional methods. In this paper, we discuss some promising applications in this direction and a number of recent works in this area; the paper also includes a general description of EC to provide a global perspective to the reader and gives some guidelines of application of EC techniques for future research.

## 1. INTRODUCTION

Researchers on Physics often work creating theoretical models to fit empirical known phenomena. These models serve to guide empirical research and also feed on the new data for generally obtain an incremental improving on accuracy of theoretical hypothesis. In many cases the parameterization of these models may be very expensive. Sometimes, small adjustments need involvement of experts knowledgeable in the problem domain. And in too many cases the knowledge they have of the problem domain is insufficient, more inaccurate or not right enough to get the expected quality results in reasonable time. In other cases, simply the dimension of the problem is such that it is impossible to cover all possibilities by conventional scanning techniques. The reader will find examples of these situations in their own area of study but to cite a current example consider the European mission GAIA planned for launch in mid-2013. The probe GAIA will be a new and important challenge that will test our technological capacity in the treatment and processing of large amounts of data [43,44] and is a perfect candidate for machine learning techniques. During 5 years of planned mission, GAIA will remain in a Lissajous orbit around the L2 Sun-Earth point getting a catalog of about a billion stars up to magnitude 20 performing up to 70 measurements of each star for determining position, distance and movement on an incredible accuracy up to 200 and at least 20 microseconds of arc. During this journey, GAIA will participate in the detection and orbital classification of dozens of thousands of extra-solar planetary systems (see 3.1.1) performing a comprehensive survey of objects ranging from a large number of Solar System minor bodies to objects of great scientific interest in the nearby universe. Probably the raw data of GAIA will take years to qualify and be processed after the end of the mission. On the following pages will discuss the virtues of evolutionary algorithms and their applicability to the study of raw data streams and will see how evolutionary computing techniques can help in the classification of large volumes of information and also in the processing of high dimensionality. We will also talk about how such techniques have been successfully applied in the recognition of characteristics and signal patterns search and in the setting of mathematical models to experimental evidence, and how evolutionary algorithms can adapt successfully to repetitive problems that need to be re-parameterized.

## 2. BRIEF RUN THROUGH EVOLUTIONARY COMPUTATION.

The first question we must answer is: I need to use Evolutionary algorithms?. To answer that question we must ask a simpler: Which are the benefits of Evolutionary algorithms over other kind of techniques? First, evolutionary computation techniques are primarily a family of multivariate optimization techniques with a high scalability, for this, are widely applied to problems of optimization with large and complex search areas (e.g. complex landscapes, noisy environments etc), recurrent and frequent scenarios in physics and especially in the field that concerns us in our study. We can to find an academic example in our star. The sun is barely eight light-minutes of distance from the Earth, since the beginnings of astronomy the study of the sun and solar flares has played an important role due to the large influence that it exerts over our planet (the sun is the principal motor of climatic changes, the source of energy that maintains photosynthesis and for that reason the base of the trophic chain, and the origin of the seasonal variations). Because of all this, a numbers of studies have been aimed to studying the sun, dynamics of heliosphere, and the forces acting on the solar corona. [1-3, 12-17]; and in spite of it, our knowledge of the same is still limited and there does not exist a complete mathematical understanding permits us to carry out trustworthy predictions about behavior of the sun or its evolution in a short or medium term.

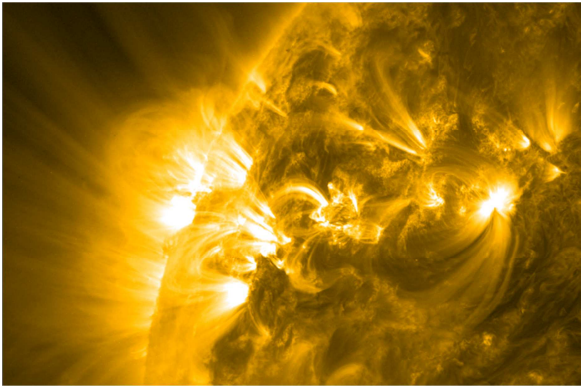

**Figure 1. A photograph of the solar surface showing its complex patterns due to convection and magnetic forces. Image: Courtesy NASA/JPL-Caltech**

For example, according to Cayrel y other recent publications [10,11] Lithium is 200 times more scarce in the sun than is predicted by the physical models, and it seems that this deficiency is also to be found in the solar-type stars near us [4,5], but the reason for this scarcity is unknown. Study of composition, evolution and stellar dynamics need of costly numerical models which feed off hundreds of variables becomes computationally unmanageable. However metaheuristic techniques such as evolutionary algorithms allow us to incorporate empirical knowledge, or obtain acceptable solutions even with a partial or incomplete knowledge of the problem. Consider the behavior of the solar corona; some authors suggest that the balance of forces acting in the solar corona can be considered as a system that tends to a minimum energy state [85]. Evolutionary algorithms can be excellent candidates to these problems, because they are very efficient to find sub-optimal solutions and to recalculate good approximations from a set of boundary conditions. A proposal in this regard can be found at Gibson et al. paper "Empirical modeling of solar corona using genetic algorithms" [86].

Problems where the deterministic exact algorithms haven't be able to find best solution or simply doesn't work tend to be the most adequate for evolutionary methods. Due to the fact that the evolutionary algorithms are heuristic, they don't ensure obtaining the global optimum on all occasions, but they normally are able to obtain a very acceptable solution in a computational time which is considerably low, as well as the fact that they don't require very specific knowledge about the problem to resolve. In order to be able to approximate these problems, in research, simulation models are used to allow validating the work developed inside realistically acceptable margins. The Evolutionary Algorithms in that environment appear as a new and innovating way of finding good adjustments and calibration parameters for mathematical models of great analytical complexity, giving also the advantage of being enormously tolerant to the necessity of re-parameterizing the model or carrying out fine adjustments when a new group of conditions appears in the system.

## 2.1 BASICS OF EVOLUTIONARY ALGORITHMS

Although not are new in concept, many researchers are unaware of the evolutionary computation techniques, or do not differentiate the existing variants. Some people use them without knowing, inside toolboxes or integrated into commercial or third party routines. Or do not exploit their potential to ignore its possibilities parameterization. Evolutionary Computation includes a group of heuristics that base their functioning on the mechanism for the natural selection proposed by biology Darwinism. A good review paper of these principles was elaborated by Goowin et al. in [19].

In evolutionary computation, a population is composed of various individuals; an individual is a candidate solution to the problem and is codified according to the necessities (e.g. it can be a binary vector); the medium where this individual develops is represented by the objective function; and the restrictions to the problem tell us just how apt that individual is to survive in that environment [18]. During searching, over these individuals (parents) of the population are applied probabilistic operators (typically crossover and mutation) to obtain new individuals (offspring of new candidate solutions) that maintain a set of properties of the ancestors which are conserved or are eliminated via a selection (deterministic o probabilistic), this process is carried out with each one of the individuals of the population until a new population is formed. This process repeats itself during a certain number of cycles (called generations in evolutionary computation) until acceptable result is achieved.

Exist various variations and improvements to this basic idea, but for simplicity, we may consider, in spite of the wide range of available different algorithms, they are all similar in their basic approach and in the usage that they give to evolutionary concepts, and differ principally in the way they represent the information and in the priority of operators applied. The algorithms of evolutionary optimization (EAs) represent the parameters to optimize inside a structure similar to natural genes and, consequently, mechanisms derived from the Darwinian ideas of natural selection and population as drift, are used. Genetic Algorithms (GAs) as a particular case, are a type of evolutionary algorithm that has proved to be very effective in the optimization of non-lineal processes [6, 7], or with noise saturation. Also, the GAs algorithms can embrace and apply themselves with success to a wide spectrum of problems and for its design it is enough to have a minimal a priori knowledge of the system. This converts the genetic algorithms into a paradigm of desirable applicability in a large number of areas where the complexity of the problem makes other types of methodologies inadvisable.

A different evolutionary approximation is the Genetic Programming (GP). This paradigm permits embracing non lineal optimization problems based on a symbolic language. The paradigm used in genetic programming also uses Darwinian selection principals such as selection based on fitness, but the genetic operators now acts over symbolic trees. For example, each one of these trees could be made up of sentences of a determined programming language. This leaves behind the conventional schema and differs from the GAs principally in what respects its representation system. The structures subjected to adaptation are generally complete and feasible programs, o hierarchical wholes of appraisable rules of dynamic form and with distinct sizes and shapes.

Despite the variety of variants and diverse paradigms in evolutionary computation [20], most authors agree in classifying the different variants as:

- Evolutionary programming
- Evolutionary strategies
- Genetic algorithms
- Genetic programming

The following paragraphs give a quick look at each of these variants.

### 2.1.1 EVOLUTIONARY PROGRAMMING

This was proposed by Fogel in [21]. In this paradigm, intelligence is viewed as an adaptive behavior. Fogel used the evolutionary programming to evolve a finite state automata, in such a way that they were capable of predicting future sequences of symbols that they were to receive. Fogel used a compensation function to indicate if automata was good or not at predicting a certain symbol. A generic algorithm for evolutionary programming method is listed as Algorithm 1.

Algorithm 1: Evolutionary Programming.

1. Randomly generate a population
2. Evaluate the aptitude of the population
3. Repeat
    4. Apply mutation operator to each individual of the population
    5. Evaluate each candidate which has resulted from the mutation
    6. Carry out the selection via roulette between parents and child
7. Until the finish condition is achieved.

In evolutionary programming no recombination operator is applied, the reason is it simulates the evolution on the level of the species, and as we know, different species can't recombine to create new individuals.

### 2.1.2 EVOLUTIONARY STRATEGIES

These were developed by Rechenberg in [22] in an attempt to resolve industrial hydrodynamics problems. The first version, called (1+1)-EE or two-member evolutionary strategy used only one parent and one child. This child self-maintained by itself if it was better than the parent. In the generation the new individuals used the function:

$$\overline{X}_{t+1} = \overline{X}_t + N(0,\sigma) \qquad (1)$$

Where $t$ refers to the current generation and $N$ is a vector of Guassian numbers with a medium of 0 and a standard deviation $\sigma$. An algorithm of the evolutionary strategies is shown in Algorithm 2.

Algorithm 2: Evolutionary Strategies

1. Randomly generate an initial population
2. Evaluate the population fitness
3. Repeat
    4. Apply mutation operator to each individual of the population
    5. Apply recombination operator
    6. Evaluate each resulting child
    7. Carry out the selection
8. Until the finish condition is achieved.

Further, Rechenberg extended the concept of population and proposed the (μ+1) – EE [23], in which there are $\mu$ parents that generate only one child, which ever replace a worse father of the population. Some years later, Schwefel introduced the use of multiple children (μ+λ) - EE and (μ,λ) - EE [24]. In the first case, in the selection offspring and parents are taken equally into consideration; in the second case only the offspring are taken into consideration in the selection.

The selection of the evolutionary strategies is deterministic, for which reason only the best individuals pass on to the next generation. The principal operator is the mutation and the recombination operator plays a secondary role and can be omitted. The evolutionary strategies simulate the evolutionary process on the level of individuals, for which reason recombination is possible.

### 2.1.3 GENETIC ALGORITHMS

These were developed by Holland in the beginnings of the 1960s in the context of machine learning [25], but weren't known about until the publication of his book in 1975 [26]. However, the genetic algorithms have been used a lot in optimization, currently becoming a very popular technique. A generic genetic algorithm schema is shown in Algorithm 3.

| Algorithm 3: Genetic Algorithm |
| --- |
| 1. Randomly generate an initial population<br>2. Evaluate the aptness of the population<br>3. Repeat<br>    4. Carry out the selection of the parents<br>    5. Apply recombination operator<br>    6. Apply mutation operator to each individual of the population<br>    7. Evaluate each resulting child<br>8. Until the stop condition is achieved. |

The genetic algorithm emphases the importance of the operator of recombination (now is the main operator) over mutation, and uses probabilistic selection. The binary codification is the most common in the genetic algorithms, and given its universality, the basic operators of a genetic algorithm are defined as of the given representation. In the terminology adopted in evolutionary computation, the binary chain that codifies a group of solutions is called chromosome. Traditionally in a genetic algorithm each segment of the chain that codifies a variable is called *gene* and the value of each chromosome position *alelo* in parallelism a biological notation.

### 2.1.4 GENETIC PROGRAMMING

Genetic programming (GP) [87] was initially considered a variant of genetic algorithms as retains the basic outline of these but differs in its internal coding and how to implement its operators. In GP, a population of solutions consists of programs that encode a given task and objective of the optimization will evolve to meet the desired task best. The evaluation of the solutions is also different because it is not based on the content of the trees as is usually done in genetic algorithms, genetic programs but are rated according to their behavior when run as a program. Programs often acquire genetic elegant solutions with a degree of subtlety not provided by the programmer.

The fitness function in a genetic algorithm typically uses an interpretation function to convert its internal codification, for example a binary vector, in the parameters of the objective functions being optimized. The algorithm manipulates the bit strings without regard to their possible interpretation. Performance functions are typically very simple, usually as simple as a conversion of a bit string to a real number. In genetic programming, the interpretation of a tree of a particular expression is carried out not according to their position but their semantics and semantic association between that and the compartments of the program can be complex. Depending on the type of semantics used can be found complex genetic operators, eg crossover operators that preserve the meaning, based on logical operations, or that keep the tree balanced consistency. The genetic programming applications are increasing in fields of aeronautics and space exploration, the design of control sequences, and autonomous navigation [88-91], and also in computer vision and detection of anomalies [94, 95].

### 2.1.5 HYBRID ALGORITHMS AND OTHERS BIO-INSPIRED ALGORITHMS

In analyzed articles we found that in a large percentage using of hybrid techniques. Using hybrid techniques we can take advantage of the qualities of the different families of methods. Many bio-inspired techniques, including evolutionary methods can are used on hybridization. Evolutionary algorithms are population-based techniques, so are complementary to other techniques such as those based on agents [81], based on pattern recognition [79-80], or making logical decisions [82]. For example, GP-Fuzzy algorithms [8] includes a population of diffuse rules / symbolic structures which are the candidates to be solutions to given problem and evolution is consider as an answer to a selective pressure induced by its relative success in the implementation of the desired conduct. Fuzzy logic, neuronal networks and evolutionary paradigms can be complementary methods for design and the implementation of intelligent systems. Neural networks, for example, are very important for plasticity in the design of adaptive systems and the recognition of shapes, signs or patterns. The reader will find interesting reading the "Neural networks in Astronomy" and the paper "Introduction: Neural networks for analysis of complex scientific data: Astronomy and geosciences" both published by Tagliaferri et al. at 2003 and 2004 [83,84].

### 2.1.5 ADVANTAGES OF THE EVOLUTIONARY ALGORITHMS

All the algorithms described above share a number of characteristics and qualities that make them unique and appropriate for use in research. Among them briefly and could highlight the qualities most interesting evolutionary computation algorithms are, among others, the following:

- They operate over all population (or group of solutions) that avoids that the search gets stuck in local optimums

- They don't require expert previous knowledge about the problem to be resolved

- They can combine with other search techniques to improve their performance

- They permit parallelization in a simple way

- They are conceptually easy to implement and use

Analyzing each of these qualities is not the object of this paper and the interested reader can be directed to other specialized papers for details on each. However there is one that makes evolutionary computation techniques of particular interest to the researcher in physics. This feature is the poblational nature of evolutive algorithms. Majority of the classic mathematical methods for multi-objective optimization operates on only one individual at a time so is necessary to execute them on various occasions in order to be able to find a group of solutions to locate to allow identifying of pareto front. Evolutionary algorithms have the advantage of working with a population (a solutions front), which permit it to generate various solutions not dominated in a single execution.

## 3. EVOLUTIONARY TECHNIQUES AS A RESEARCH TOOL IN ASTRONOMY AND ASTROPHYSICS.

The number of papers that make use of bio-inspired techniques in general and particularly of evolutionary techniques has had an exponential growth in the last 15 years. In particular, the use of these techniques in astronomy and astrophysics has developed dramatically. Figure 3 may be the number of papers found in the consulted sources and increase to reach more than 150 articles published for last year.

In next sections, we conducted a survey of the number and topics of articles published and a detailed analysis of some of them. The articles analyzed are representative of a wide range of areas of research which highlights the high applicability of this family of algorithms in this field. You can see a detailed topics classification on figure 2.

As the study of each of these topics separately no sense in the context of our study we decided to reduce the number of subject areas to four, within which can be grouped most of the records found:

- Study of distant bodies for optical and radio band
- Solar dynamics, helioseismology and stellar evolution
- Study of galaxies and other supermassive bodies.
- Monitoring and study of high energy events and associated theory

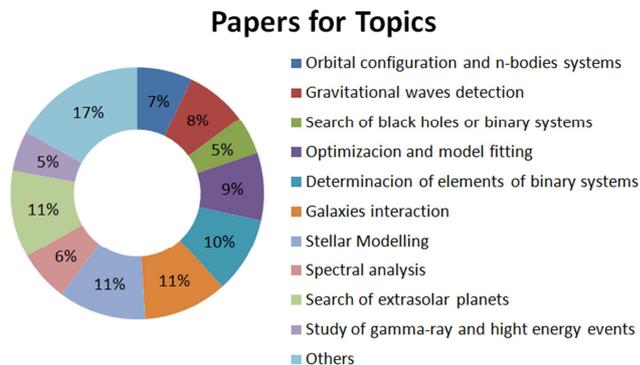

**Figure 2. Grouping of topics according to the most significant thematic clusters**

The percentages shown are virtually unchanged (or show no significant changes) throughout the study period except, perhaps, the most cutting-edge areas as the study of extra-solar planetary systems or search for gravitational waves, which barely existed at the beginning of the period but have shown a big development in the last decade.

In the following subsections we will develop each of the described topics, highlighting the more relevant works and that we believe are the more important contributions of the authors.

### 3.1 EVOLUTIVE ALGORITHMS ON USE OF OBSERVATIONAL ANOMALIES ON OPTICAL AND RADIO ASTRONOMY FOR STUDY OF DISTANT BODIES.

It is possible to consider the moment of the popularization of radio-astronomy at the end of the forties as a great point of inflection in the comprehension of the stellar systems, comparable in magnitude to the discovery of astronomical spectroscopy at the beginnings of the century because it allowed us for first time to explore the universe without limitations of the visible spectrum and consequently, opened the gate to development of radio-interferometry and the discovery or confirmation of dozens of new stellar objects types that changed our understanding of physics: black holes, quasars, neutron stars, or magnetars [27].

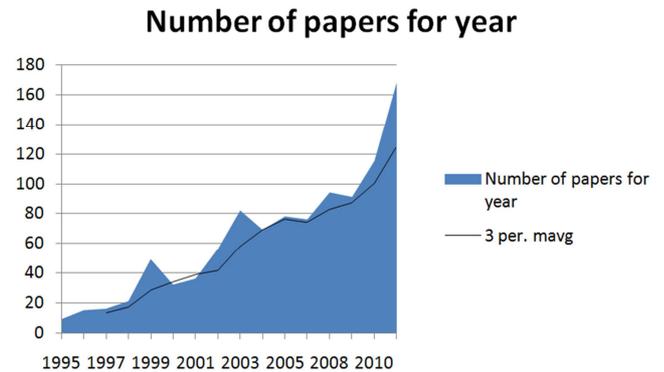

**Figure 3. Papers on A&A related areas using evolutionary computation techniques. Appearance of trend line and moving averages.**

However, only the development of the space age and the unfolding of scientific satellites permitted increasing the rhythm of discoveries, bringing volumes of information unknown before to any scientific branch. Are many the fields that open given this perspective for artificial intelligence and the number of applications increases as new and more sophisticated instruments are launched. Among these instruments we could cite the space telescope Kepler [28] launched in 2009, capable of observing simultaneously 150.000 stars in search of disturbances that indicate the existence of extra-solar planets, the Gamma-ray Large Area Space Telescope (Fermi) launched in June 2008 and that has been for two years tracing a map of gamma radiation sources that will permit studying the evolution of the galaxies and the history of its formation, and the future Laser Interferometer Space Antenna (LISA) [30] which will allow detection of the presence of gravitational waves and will help to validate one of the most controversial aspect of the general theory of relativity (see 3.4) in which the evolutionary algorithms are already having an important role working with databases of simulated sensor data that allow making the mission profitable from the first moment and guaranteeing the obtaining of satisfactory information.

One of the research areas that most publications have generated in recent years in is the search for planets with potential for life beyond our solar system. After the discovery of the first exo-planet orbiting the star 51-Pegasi b in 1995 [31], the pace of discoveries and the techniques that permit indirect detection of these types of bodies have increased exponentially to reach the 61 planets discovered in the year 2009 and a total in 2010 of 466 planets. With the launching of the European mission Corot [29] and the American space Telescope Kepler, mentioned above, missions specialized on search for exo-planets using the transit method is hoped that the number of discoveries will considerably increase over the next 5 years. The planetary transits are very short and rarely visible from Earth, besides the possibility of detecting them is inversely proportional to the axis of planetary orbit and directly proportional to the time of continuous observation. So this technique is really effective when you monitor the number of stars simultaneously is very large (eg. Kepler mission uses photometers capable of detecting variations in the shining of 150,000 stars simultaneously and up to 20 parts per million). In 2010, Chwatal et al. successfully applied evolution strategy algorithm to the problem of detecting planetary transits in systems from multiple time-series data from Corot and Kepler [96] (as described previously both observatories use the transit method). A simplified formula to calculate the approximate duration of the transit of an exo-planet is showed as equation 2. In formula given, $d*$ is the diameter of the star expressed in solar radii, $a$ is the axis of the planet's orbit in astronomical units and $m*$ is the mass of the star measured in solar masses. As you can see the transit times are very small compared to the planet's orbital period. This means that there are other more direct methods for their detection. One of the most popular methods for the discovery of extra-solar planets has been the method based on radial velocity ($RV$). This method bases its functioning on the gravitational influence, that although small, the planetary body exert on its star. In 1997, Lazio et al. used a basic genetic algorithm to fit Keplerian orbits and to explain the irregularities in the periods of some pulsar by the presence of other secondary bodies [60]. They observed the genetic algorithm was more efficient and found better solutions than the traditional exact methods and the new precision allowing the discovery of a planetary companion to the pulsar PSR B0329 +54 and identified two other potential candidates: B1911-04 and B1929 +10. In the stars, little orbit anomalies as influence of planets can be measured via the Doppler shift in the light spectrum that can be measured and to reveal the presence of a planet, two planets, or a complete planetary system.

$$\tau_C = 13d* \sqrt{a/m*} \cong 13\sqrt{a} * hrs \qquad (2)$$

Mainly the complexity of this process is to detect the fluctuations that the planetary system altogether provokes on the orbit of the star, which can lead to months of careful observations, for which reason the model should start from the supposition of a system and find the levels in concordance with the data observed based on a variable number of possible planets, that also do not have to have similar orbital configurations, since there could exist stable systems made up of planets with distinct orbital inclinations, with distinct eccentricity, from almost circular orbits to orbits as elliptical as those of the comets, with distinct masses, and densities, with moons or without them, and which orbit their star at distinct velocities. This variety of situations makes traditional analysis, as based on Fourier's transformed events, that tries to split the original signal in a series of signals that in total explain that the wave observed (minimizing residual noise) is not correct due mostly to the fact that in this approach the orbits need to be approximated as circular orbits, which has the result that idealized adjustments are reached. On the contrary, from the point of view of an evolutionary algorithm it proves easy to explore the group of possible solutions and propose tentative solutions with relative rapidity. It is for this reason that these types of techniques are applied each time more assiduity to the analysis of observational raw data. In 2010, Rozenkiewicz y Gozdziewski [32] proposed a hybrid algorithm using a genetic algorithm and a simplex algorithm [33] to study star HD240210 which showed abnormalities that were not explained in the 1-planet keplerian model but could be interpreted in context of the existence of an additional planet. The genetic algorithm was used to explore possible variations for orbital parameters like orbital eccentricities, orbital periods, and time of periastron passage and eccentricity. The algorithm prioritized solutions that produced best-fit model parameters for a 2:1 mean motion resonance configuration. The results of experiments ended in the discovery and subsequent confirmation of planet HD240210b.

In the study of other more massive bodies, like galaxies, already at 1995, Charbonneau et al. proposed an open framework that would allow integration of different techniques, including several evolutionary techniques in field work and research in physical sciences. These efforts resulted in the software PIKAIA [34] which was successfully used in the study of the solar core rotation and galaxies interaction. It would be the basis for many subsequent studies. Also in 1995, Lazio and Cordes, proposed a possible method for searching of partner bodies and possible planets around pulsars of known frequency by using a genetic algorithm to explore possible systems within a given acceptance and margins basing on the manner those secondary bodies could affect to pulses creating small but measurable perturbations in their pulsation frequency [35]. At 2008, Chwatal et al. carried out other particularly interesting work [36], in which they proposed the use of evolution strategies for the characterization of extra-solar planets for detection of planets inside its habitability zones.

Evolution strategies are considered by many researchers [38, 39] as more effective solving optimization problems with continuous parameters that genetic algorithms. This is mainly due to limitations in the way of encoding the AGs and the absence of self-adaptive mechanisms to allow exploring the fitness landscape structure. As we described earlier, in Evolution Strategies algorithms, mutation and not crossover is considered the main operator, while the selection is usually performed deterministically. In the classic process of mutation, the mutations may affect one or several parameters of one or more individuals of the same generation. In the algorithm proposed by Chwatal the mutation operator is forced to act for the subset of parameters for a single planet in the system each time. This type of mutation imitates the evolution of planetary systems in real interactions. In view to improve the possibilities of convergence of the algorithm, they limits the solution space to disallow unstable systems considered unstable by the Hill stability criterion which we can calculate the minimum radius that should keep the orbits of two planets between them to ensure that no destabilizing to each other and therefore end up crashing.

| Name | R/RE | Orbital period | S.major axis | Estimated surface temperature | Spectral type |
|---|---|---|---|---|---|
| KOI 701.03 | 1,73 | 122,4 days | 0,45 UA | 262 K (-11º C) | K type (4869K) |
| KOI 1026.01 | 1,77 | 94 days | 0,33 UA | 242 K (-31º C) | M type (3802K) |
| KOI 854.01 | 1,91 | 56,05 days | 0,22 UA | 248 K (-25º C) | M type (3743K) |
| KOI 268.01 | 1,75 | 110,37 days | 0,41 UA | 295 K (22º C) | K type (4808K) |
| KOI 70.03 | 1,96 | 77,61 days | 0,35 UA | 333 K (60º C) | G type (5342K) |

**Table 1. Some planet candidates similar to the Earth discovered by Kepler. Currently Kepler team recognizes error range for size is 25-33%. And the error in the values of temperatures could approach 20%.**

Also is contemplated a special situation according to which, if a planet is on the edge of his Hill's area can jump to another different orbit creating a sort of tunnel effect that will be very interesting concept for escaping from local minima. For example, we can suppose the case in that most of the planetary orbits parameters have been determined but the final position of a planet that is difficult to determine, and is incorrect even in the model. In this circumstance, this mechanism can be very useful to allow the planet to reach configurations that are more responsive to the model. As an additional mechanism to accelerate the convergence preventing the algorithm falling on solutions that are even close to being valid no have value because it does not fully represent the sampled data. They also define a new mechanism named as *evolution path* which consist on establish a maximum time of survival for each family of solutions. The way of evolution path is to encourage that if a set of solutions that have exhausted their survival time then the planet's worst fit is removed and replaced by a new one that occupies a random orbit consistent with the system, thereby increasing the ability of the algorithm to explore new areas. To perform a realistic comparison the authors used two previous papers referring to v-Andromedae and to 55-Cancri systems [40, 41], both reporting discoveries contrasted using conventional methods, obtaining the same or better planetary configurations.

In the same line, Konacki and Góździewski proposed in [61] a modified genetic algorithm that integrates the checks of stability as inner fitting procedure. Calculation of this stability criterion is based on estimation of Lyapunov Exponent for a Hamiltonian, this indicator uses concepts from theory of attractors and their behavior is related to the measure of chaoticity of a system. The algorithm described, named GAMP (genetic algorithm with MEGNO penalty) was shown particularly good at finding planetary configurations in low-order mean motion resonances.

Another interesting variation is proposed in the paper of 2008 signed by Attia et al. [77] which in this case proposes a genetic algorithm with dynamic population adjustment to fit orbital parameters of the invisible star in the binary system η-Bootis basing on the variations in radial velocity for visible component observed by Moore in 1905 and confirmed by Bertiau in 1957 [42]. Datasets from different observations was unified and normalized resulting in an optimization problem of moderate difficulty. The orbits of the system elements were characterized by 6 parameters: $p$, the orbital period, $\tau$, time of perihelion passage, $\psi$, the longitude of perihelion, $e$, the orbital eccentricity, $K$, the orbital velocity amplitude and $Vo$, the system's radial velocity. The proposed algorithm associate crossover and mutation rates with each chromosome in every generation depending on the fitness value of this chromosome. This technique is so called feedback capability and was introduced in 1994 by Srinivas and Patnaik [43]. The algorithm starts by performing a wide search which allows a large range of values for the parameters considered only bounded by the observational constraints to locate the area of the fitness landscape that is more promising for, once located, perform a search of fine tuning that permits to locate the optimum. The results were compared with those obtained in the same system [44], achieving greater precision in the value of the parameters which have repercussions on higher quality adjustment and a computational time seven times lower than conventional methods.

3.2 EVOLUTIVE ALGORITHMS ON STUDY OF SOLAR DYNAMIC AND HELIOSISMOLOGY

The study of interacting forces inside stars and stellar evolution is fundamental to our understanding of the universe. The physics of stars is inferred through observation and theoretical construction, and development of complex mathematical models of its internal structure [92, 46, 65]. On simple terms, stars genesis occurs in regions that are dense in gas and dust as nebulae. When destabilized, some parts of the cloud may collapse under the influence of the gravity, to form a proto-star. If the agglomeration is sufficiently dense and hot, the core starts a nuclear fusion, creating a main sequence star. The characteristics of the resulting star depend primarily on its initial mass. The more massive the star will become more luminous and therefore faster exhaust the hydrogen fuel at its core, although this also depends on the type of fusion reaction that predominates in the star [93]. During this time the star produces energy by hydrogen fusion, a process known as proton-proton reaction, the most important cycle in terms of energy production in solar-type stars. The different paths of stellar evolution known representing the most important star types are listed in the actually accepted Hertzsprung-Russell diagram drew in Figure 4 showing evolutionary path different colors for the four types of stellar masses more representative.

The most direct way to study a star is by studying its light; Across the spectrum of a star is not only possible to determine the chemical composition of the surface layers, but also some characteristic parameters of the star, for example: The speed of rotation of a star can be determined by observing the absorption lines sharpness; where the most defined corresponds at lower speeds, and more blurred higher speeds. When the spectral lines are widened with pressure is indicative of more frequent atomic collisions in a dense gas become diffusing certain energy levels. And when a spectrum is split absorption lines indicates the presence of strong magnetic fields (this phenomenon is known as the Zeeman Effect).

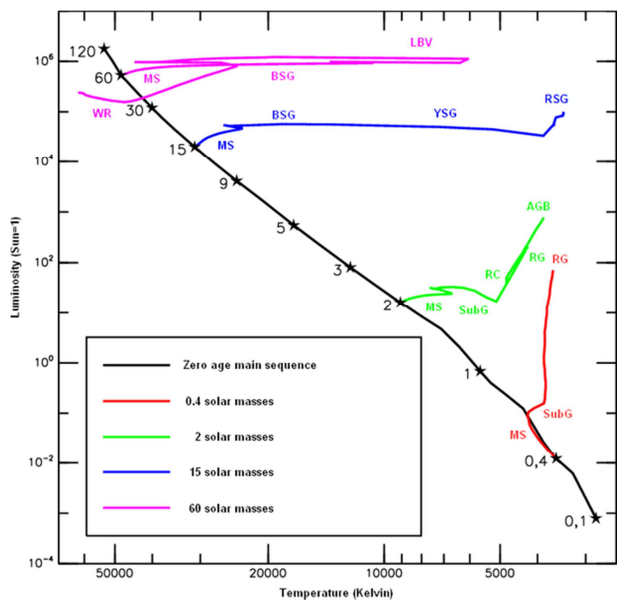

**Figure 4. Same stellar evolutionary tracks for single stars, zero initial rotational velocity, and solar metallicity: AGS Asymptotic Giant Branch, RG Red Giant, SubG Subgiant, MS Main Sequence, RC Red Clump, BSG Blue Supergiant, YSG Yellow Supergiant RSG Red Supergiant, WR Wolf-Rayet stars, LBV Luminous blue variables. Image: Wikimedia Commons, reproducible under the Creative Commons Attribution/Share-Alike License**

The first work that we find where evolutionary techniques applied to the analysis of stellar spectra dates from 1999 and is signed by Metcalfe [50]. It proposed using a genetic algorithm to infer the characteristics for binary stellar systems by adjusting the observed light curves with those generated by theoretical models as proposed by Wilson-Devinney [48] and Stagg and Milone [49]. Due to the computational weight of calculations they had to implement a distributed infrastructure on a grid of twenty-five workstations that will deal with the simulations commanded by a central server where the genetic algorithm will to organize the population and take control of work assignments. The proposed algorithm began generating a set of 1000 arrays of randomly distributed parameters which are sent to the slave nodes to be responsible for calculating the theoretical curves according to the model UBV [51] (The letters U, B, and V stand for ultraviolet, blue, and visual magnitudes, which are measured for a star in order to classify it in the UBV system.); after receiving the results the master node sent a new job to slave node and uses the calculated curve to measure the differences between the data and the actual data observed. The fitness value for all parameter sets are assigned based on the magnitude of the observed differences. As expected they found that the performance of the algorithm is not significantly decreased with increasing number of parameters, allowing a greater number of parameterized variables exceeding expectations at the beginning of the study.

A little time earlier, in 1996, Gibson and Charbonneau [58] had successfully applied a genetic algorithm to solar coronal modeling, an area particularly interesting consequence of the nonconformity appreciated in actual measurements with respect to the theoretical measures.

In 2000, Jagielski [99] proposed a genetic programming algorithm for prediction of solar flares, based on predicted series generated from the information on sunspots and solar flares available since 1850. The data set was divided into two parts, 10% of the series is split into "training sets of 200 records," while the rest was reserved for testing the quality of the predictions. The results obtained exceeded the percentages obtained in previous studies using neural networks [97, 98]. Currently and under the European program MIERS (Mitigation of Ionospheric Effects on Radio Systems), genetic programming algorithms are used along with other prediction and forecasting methods like neuro-fuzzy predictors to anticipate solar flares that could damage satellites and critical communications systems through sophisticated models Ionosphere geomagnetic activity [100].

In 2002, Metcalfe and Charbonneau publish a possible implementation for a distributed genetic algorithm for determining the globally optimal parameters for a mathematical model used to obtain physical and structural information about the stars through the observation of their oscillation frequencies [78]. The mathematical model was adjusted over a dataset that was obtained from sampling the behavior of a white dwarf star getting interesting physical results and providing insights into the history of the nuclear reactions that took place in the star getting estimations about the amount of carbon that has been formed on their core, or if have formed heavier elements like oxygen, and other parameters allowing to characterize the star and know the moment of their lifecycle. Although less ambitious, this work has a clear precedent in the paper of Tomezyk et al. of 1995 [59] considering of the application of genetic algorithms in the field of helio-seismology.

Helio-seismology is especially important on the study and determination of stellar evolution history. For example, solar-type stars finish their lives when the hydrogen, which maintains its primary nuclear cycles, begins to run low. At this time the helium that has been produced during the life of the star begins to contract by the action of gravity increasing the rate of burning hydrogen in the layers surrounding the core, converting the star a red giant several times greater than the original star (see Figure 4, green evolution path). While the helium core continues to compress further under the gravitational influence of its own mass, temperatures and densities reached start getting the fusion of helium into heavier elements like carbon. This carbon in turn can combine with two atoms to form helium and oxygen, and larger stars even heavier elements. At the end of the life of the red giant phase now known as Asymptotic Giant Branch (marked by red line on figure), these elements are dispersed to form a planetary nebula and in the center of the system is a hot white dwarf. In this type of stars already nuclear reactions do not occur so its brightness is due to its high temperature and decays while the star loses heat to space. The way it produces this cooling process can create gravity waves and to do vibrate to the star periodically in a certain manner; in this case we talk about of pulsating white dwarf stars. These vibrations send waves into the depths of the star in the form of seismic waves and earthquakes turn these small changes in brightness appreciably (only about 1%). Using high-speed photometric detectors, researchers can look closely at these changes and build computer models that reveal the internal structure of the white dwarf and its nuclear history. To know more about the physical processes that occur in the formation of planetary nebulae can be consulted [51, 52]. They are also a valuable source of information over the subject the papers of S. Turck-Chieze, W. Dapper and teams in particle physics and the standard solar model [53].

Currently are known 26 types of variable white dwarf stars, these stars are called in the literature ZZ Ceti stardue to the first star of its kind was discovered in the Cetus constellation. Usually, these stars have a carbon-oxygen core surrounded by a wrapper of pure helium, and over it floats a smaller layer of hydrogen. The pulses emitted by ZZ Ceti stars are non-radial pulse multi-period, with periods between 100 seconds and 30 minutes. A radial pulse when the star is oscillating around an equilibrium state by changing its radius while maintaining its spherical shape. The radial pulse is only a special case of non-radial pulsation. Non-radial pulsation occurs when parts of the star's surface feel pushed in one direction while other parties do so in a different direction at the same time. The algorithm they proposed was a parallel genetic algorithm based on that developed by Charbonneau in 1995 [34] that operates within a client-server architecture. It would really be considered questionable whether the algorithm implemented as a parallel genetic algorithm since it is not taking advantage implied the use of parallelism in these kinds of algorithms [54] but used this as a resource against the high computational cost of constructing and simulation of different models. The algorithm is running in a Linux cluster composed by a variable number of nodes; the server node running the master program that initialize a client application on each of the slave nodes. The master program creates one parameter vector for each node (initially content is random) and generates with them a population of candidate solutions, then the server sends one of these candidates to each client to be evaluated its potential as a solution. The client manages information about mass, temperature, composition and structure inside its parameters vector and initializes a simulation model that will advance the state of the star so far measured. Based on resultations algorithm calculate the pulsation periods should occur in the star following the currently accepted model described by Kawaler and Bradley in 1994 [55-57]. Obtaining the observed data not only provides valuable information but that indirectly involves the validation of theoretical models.

An interesting contribution is the proposal by Ordonez and Dafonte in 2010 [69]. In this paper the authors propose a hybrid algorithm that uses a genetic algorithm to adjust the recognition fee of a neural network that handles the classification and characterization of stellar mass objects by studying their emission spectrum. The experiments were developed from a simulated dataset of 9048 samples in wavelengths between 847.58nm and 873.59nm consistent with the data expected to be generated by the satellite GAIA RVS instrument. It is expected that data from GAIA inevitably contain noise of different nature (level of sensor sensitivity, background noise, and instrumental noise, etc). Thus is introduced artificial noise in the samples at different levels. The genetic algorithm allows efficient handling of this dirty information and also work well with the high dimensionality of the data selecting the most relevant features of the spectrum. The genetic algorithm utilized is quite simple, applies regular cross-over strategy and algorithm selection classic roulette wheel and as in previous cases elitist behavior is introduced. A novel idea is the way of choosing the objective function; in this case it comes to optimize the recognition capability of the neuronal network. Therefore, the objective function is based on the accuracy achieved by the network for a correct classification. For this end, the training method, reserve 30% of the data for the configuration phase, so that the remaining 70% of the samples will be those really destined achieving results. Thus, the genetic algorithm is responsible for maximizing the use of the learning phase, significantly improving the effective capacity of the network classification, demonstrating the capacity of evolutionary algorithms to work with other techniques into hybrid solutions to improve the results of the two techniques separately.

### 4.2.3 EVOLUTIVE ALGORITHMS ON STUDY OF GRAVITATIONAL INTERACTIONS AND PHYSICS OF GALAXIES AND OTHER SUPERMASSIVE BODIES.

In the early universe, star formation rate was higher than today, frequently colliding galaxies, generating a wealth of new stars with each encounter. Although these events are now relatively uncommon, galactic interactions continue to forge stars and galaxies shaping. The study of galaxies is similar to the study of continental plates on Earth. When dealing with time scales much greater than the length of a human life is impossible to empirically observe the event. However, the dynamics of interactions between galaxies and deformation characteristics and ways that these interactions act can to give a lot of information and allow us a projection through time. However, these simulations typically require many hours of computer and need huge precisions due to the size of physical magnitudes implied.

Theis et al. in 1999 [70] used a modified version of the Charbonneau algorithm of 1995 [34] to study the evolutionary history of the galaxy NGC4449, a dwarf galaxy of Magellanic type which is currently very active star formation through its gravitational interaction with DDO125 other nearby dwarf galaxy. Mathematically the proposed problem can approach like the optimization problem of fitting an N-body simulation for a given observed dataset. The precision of these measurements depends largely on the increase in accessible particle number. The individuals in the population encode a set of fitting parameters which will input values for simulations. Among the parameters considered are the encounter eccentricity, the mass ratio, the orientation of the orbital plane, the inclination axis, and the position of the galactic disks. The quality of individuals is measured using a quality function based on the similarity between the data obtained by simulation and observed data where unstable parameter sets or impossible configurations are penalized. They used roulette-wheel selection and uniform crossover, which together with mutation mechanism are responsible for generating variability. Introduced mutation rate is relatively high, 0.5%, and the algorithm is responsible for adapting during the course of the search to avoid inbreeding (homogenization of gene pool by one or a few dominant genes). During testing was found that the genetic algorithm typically converged after 100 generations, which is about $10^4$ simulations. And while for similar results using a traditional model 3CPU GRAPE would have needed 3.4 years, the presented model reduced the CPU requirement to 5.6 hours on a 150MHz-Sparc10, which represents a difference large enough to make a problem manageable previously unmanageable. In the same line, in 1997, Wahde et al. proposed the application of a genetic algorithm to obtain scientific information of interest for the characterization of galaxies. They take same characteristical values like the scale radii, scale heights, disc inclinations, velocity dispersions, or Masses ratio from real observations if galaxies on interaction where had parameters unknown and some measures been strongly affected by noise [71]. The final results were very good despite the was not much innovation in the type of algorithm used, but so in the way of choosing the most promising solutions that now considers the existence of certain characteristic shapes such as bars, rings, ovals or spirals, thereby introducing much more specific knowledge and improving the quality of the results.

In a more recently paper, in 2001, Wahde et al. [62] used a genetic algorithm to study the influence of past interaction between and the galaxy NGC5195 and the Messier M51 (commonly known as the Whirlpool Galaxy). On this occasion, the information encoded on chromosomes up to 11 parameters of both galaxies, as their masses, their position relative to the observation points, their radial velocities, direction of their axes of rotation or inclination and position angles. In this case the novelty of this study is how to perform the simulations. The simulation model does not focus on highlighting the interactions in a short period of time but that the positions of both systems are integrated from beginning to influence each other as point particles until it reaches the time of observation, trying to understand the mechanisms of the interactions between. During testing, the algorithm works with samples of 5000 particles in each simulation and the simulated time period exceeds 1.05 Gyr (1000 time units). The population size was 50 individuals per generation and the algorithm seemed to converge on average in generation 70. Although a second phase of the experiments got a greater accuracy result using populations of 250 individuals, where the algorithm converged after 200 generations. What is interesting here was that the increase in the time it takes the algorithm to find a solution within the established acceptance margins did not grow the way you would expect from a traditional algorithm to grow the number of points. Two years later, in 2003, Theis and Spinneker created the first high resolution intensity and velocity maps for the M51 system based on these data where in-depth study of the asymmetry of the spiral arms of this galaxy [63]. In same year, Teuben proposed an algorithm based on genetic programming [102] to approximate the dynamics of galactic interaction using the toolbox based NEMO (Stellar Dynamics Toolbox) [101]. Later, in 2004, Li, Yao and Frayn, proposed an evolutionary approach using genetic programming to mathematical modeling of the brightness distribution in elliptical galaxies [103]. In this case using an encoding based on mathematical operators. We also consider especially remarkable work carried out by Cantu and Kamath in 2002 [64] which propose a hybrid solution AG-ANN, consisting of a genetic algorithm and back-propagation neural network with one hidden layer. In this paper, genetic algorithm performs the task of finding the best weights vector, and finds also the smallest number of input features to allow the neural network correctly classified observations of distant galaxies. They used a genetic algorithm with a population of 50 individuals, binary encoding, pair-wise tournament selection and multipoint crossover. Although this simple model, algorithm reached in some configurations accuracy above 72.69 percent. Again the hybrid solution results exceeded expectations and were better than expected. Neural networks are often used in hybrid solutions as a substitute for genetic programming algorithms. In this case the knowledge is lost less specific encoding is stored in the network connections.

### 4.2.4 EVOLUTIVE ALGORITHMS ON SUPERNOVAE STUDY AND OTHERS HIGH-ENERGY EVENTS REGISTRY.

The fate of a star depends primarily on its mass, the stars of mass greater than eight times the Sun usually finish their lives in a core-collapse supernova, while the smaller stars, probably end up forming a planetary nebula, or become white dwarfs [104]. In turn, the remnant of a supernova may have different endings depending on its mass [45- 47]. If the initial mass of the star is greater than about 30 solar masses (the exact limit depends of the star metallicity), part of the outer layers can't escape of the gravitational pull of the neutron star and fall on it for causing a second collapse and creating a black hole as the final remnant (marked by blue and purple lines on figure 4). Two close binary stars can follow more complex evolutionary paths, such as mass transfer onto a white dwarf companion that can potentially cause a supernova. Some types of supernova as the hypernovae is often associated with one or more shoots of high-energy emissions generally in the form of gamma rays (and also other lower-frequency emissions in X-rays, ultraviolet and visible light). These emissions are very important in the field cosmological because they are detectable at huge distances. For example, the detection of a gamma-ray burst to 13 billion light years away by the NASA SWIFT observatory in 2009 allowed to confirm the farthest supernova explosion recorded. This event lasted only a few seconds was named GRB 090423.

But gamma-ray bursts are not something exclusive to the collapse of stars but also may to reveal many of the most violent events in the universe for example interactions of objects which have huge mass, this is the case of super-massive black holes or by interactions in binary systems consisting of two black holes or a black hole and a neutron star. This is the assuming that Zwart et al. studied in 2001[72]. In this work they used an evolutionary algorithm for fitting a set of numerical models to several real observations of gamma emissions and tried to explain these by generation of high-energy emission jets in a rotation system where particles of ray interacts with the interstellar medium. His model needed a total of 14 parameters, so the algorithm chosen follows a scheme where each chromosome is formed by a vector containing each of these 14 real values. The selected mutation rate was very high for this kind of algorithms, over 5%, allowing exploratory high capacity although it might not be appropriate in some cases. Both, the mutation as in the crossover operators parameters are monitored and they are maintained within margins of acceptance established based on observed data and the limitations of physical variables and criterions to avoid occurrence of very similar solutions (within certain ranges of similarity). In this manner, variability in the offspring was forced to obtain a greater explorative capacity. Crossover was performed in a uniform manner, so that a gene belongs to a parent has a 50% chance to spread her son but the paper does not detail how recombination occurs and whether this will retain the solutions with or without physical sense.

Described scenery is one of many situations where gamma emissions can occur, some this sceneries can product another event of equal scientific interest; it is the generation of gravitational waves. Gravitational waves are an untested theoretical result of the theory of general relativity according to which an extremely massive rotating body should create a distortion in the form of space in a manner similar to the waves as they move on a fluid. These gravitational distortions although very weak and undetectable from the ground would be measurable by high-precision interferometers. To accomplish this task, in 2004 the United States lunch to orbit the observatory LIGO (Laser Interferometer Gravitational Wave Observatory). Many papers have been proposed to suggest the use of different techniques of data mining to study of LIGO data streams. One of the most interesting is the proposal of Lightman et al. at 2006 [73] which proposes the implementation of an genetic programming algorithm to maximize the chances of detecting gravitational waves minimizing the risk of false alarms in LIGO. In Lightman algorithm a chromosomes encode a tree structure where each node is formed by logical rules that represent a logical clause and each branch forms a decision path that allows to detection algorithm to deduce whether affirmative or negative cases. The fitness of a candidate is given by its ability to generate a correct detection. Result of the selected encoding the crossover operator has to work at the logical level. Given to both parents, the algorithm creates new sub-tree, choosing at random points in the parent trees that create a logical partition clause, generating the new offspring as the union of two sub-trees retain a logical coherence. The operation of mutation, consist in this case to change a logical operator in the clause (in the simplest case) or the introduction of a new full sub-tree (in the case more complex). The fitness function takes in consideration the penalty for false positives and other characteristics like the tree size or the difficulty of processing. The results suggest that the evolutionary program proposed helped to create effective solutions potentially valid for the detection of gravitational waves from the LIGO raw datasets and identify unitary building blocks in these solutions that could help improve existing detection algorithms and discover new creative solutions.

The LIGO mission is a natural successor in the probe LISA. This mission seems much more interesting for future studies. First, as a consequence of its higher resolution and for larger volume of data generated and secondly, by its greater ability of this study the origin of the detection objects. The first paper that made use of genetic algorithms applied to the analysis of data from LISA was introduced by Crowder and Cornish in 2006 [74]. It presents an effective method to isolate potential sources of gravitational waves from the possible tens of thousands of overlapping signals in the LISA sensors data stream, reducing the level of confusing noise. The evolutionary approach is used as a first step in cleaning up the signal and constraining the parameters and allowing a further refining process using a Markov Chain Monte Carlo algorithm. Each candidate solution represents a waveform template and within each individual a chromosome is formed by set of the parameters that define the shape of the signal. Again is used uniform crossover operator witch to take the center point of array parameters as partition point, and recombination is thus carried by one half from each parent. More interesting is the mutation operation used; In the article the authors present a comprehensive empirical study where they test different mutation rates and their influence on the speed and quality of convergence. Also is weighted the influence of parameters such as the existence of elitism and population growth. The study proposes an improved algorithm where these parameters are analyzed as to optimize parameters allowing it to evolve and adapt dynamically. For example, they take greater probability of mutation in the early stages of the search (about 0.4) and much lower (less than 0.02) when the convergence is close. In a similar research, in 2009, Petiteau et al. [76] presented several improvements to this algorithm by introducing some interesting ideas, for example, the concept of temperature which is used in other techniques as simulated annealing to dynamically adjust the operating capability of the algorithm. Or the concept of brother, to explore new areas based on promising areas already explored. And two closely related concepts, the local mutation and freeze bits, both allowing fine adjustment in advanced stages and a better control over the premature convergence. The genetic programming algorithms can also be useful in detecting high-energy events and in particle characterization. In 2005, a multidisciplinary team of scientists from the U.S., South America, Europe and Korea [106] proposed a genetic programming algorithm that showed promising results in detecting and predicting trajectories of high energy particles within the FOCUS project [105]. Again in this case the encoding used by the mathematical algorithm is a tree where each node represents a simple math (arithmetic, trigonometric, logical), and each branch a variable kinematic or a characterization parameter of the particle.

Finally we consider interesting to mention the extensive paper published by Gair and Porter in 2010 [75]. They investigated new methods of collaborating mixture techniques in which a genetic algorithm, similar to that described in previous papers, formed by a nested sampling algorithm [66,67] and a Metropolis-Hasting method [68] (a more efficient variant of the family of Monte Carlo Markov Chain methods) were used together in a collaborative manner. In this approach which could be classified as a mixed hybrid algorithm, the genetic algorithm is responsible for guiding the search, locating areas of interest within which there will be a competitive process to form islands of local search. In this case the fitness function is the Fisher Matrix related with estimation errors. The crossover operation also takes place within each island so that the best solution of the group and the second best solution will give rise to new offspring. Tests from the mock LISA data set behaved extremely reliable manner and required only small execution times even on computers of modest benefit, besides allowing the simultaneous detection of several sources.

## 4. CONCLUSION, RESEARCH IMPLICATIONS AND LIMITATIONS

Although the application of evolutionary techniques in some fields as astronomical sciences is still incipient, these techniques are beginning to show its potential as a powerful analysis, configuration, and optimization tool in complex search space problems not treatable by conventional methods or in noisy situations, overlap or signal degradation. The combined use of these techniques together others specialized in clustering or classification issues allows obtaining of complex patterns and extracting valuable information from raw data. Its great adaptability and tolerance to the erroneous data also make it good tools for optimization and parameter fitting of complex functions as common in empirical science. An added benefit over other techniques of mathematical optimization evolutionary algorithms allow obtaining many multiple solutions, quasi-optima solutions, approximate, or in time bounded, and an easy way to avoid the dreaded local minima in complex landscapes. Perhaps the biggest limitation of these techniques is, that, as with all heuristics, in an evolutionary algorithm is vitally important to choose a correct guiding function and give to algorithm enough specific knowledge of the problem to achieve a successful convergence in computational time. Despite this, a GA does not need to make any assumptions about how the search space, hence its wide applicability.

## 5. SUGGESTIONS FOR FURTHER RESEARCH

The rapid adoption of new techniques is closely related to these benefits. In this paper we have seen that these benefits exist and future research undoubtedly find new ways and areas of applicability where initial studies show promising results. The fields of research are many, from data fitting and parameterization of models, where these techniques have shown interesting features such as to be scalable (performance does not worsen significantly by increasing the number of dimensions), its insensitivity to corruption and noise, or its tolerance to local minima, until data mining applications, search for components in n-dimensional data, training of machine learning algorithms, or as efficient alternative to analytical optimization methods of intensive CPU usage. Future papers should evaluate these possibilities and to incorporate new advances in algorithms to the fields of applied research that are already known.

## 6. ACKNOWLEDGEMENTS

Thanks to Dr. Alberto Castro-Tirado for their corrections and suggestions to this paper.

## References

[1]   The Current State of Solar Modeling, Christensen-Dalsgaard J, Dappen W, Ajukov SV, Anderson ER, Antia HM, Basu S, Baturin VA, Berthomieu G,

[1] Chaboyer B, Chitre SM, Cox AN, Demarque P, Donatowicz J, Dziembowski WA, Gabriel M, Gough DO, Guenther DB, Guzik JA, Harvey JW, Hill F, Houdek G, Iglesias CA, Kosovichev AG, Leibacher JW, Morel P, Proffitt CR, Provost J, Reiter J, Rhodes EJ Jr, Rogers FJ, Roxburgh IW, Thompson MJ, Ulrich RK. Science. 1996 May 31;272(5266):1286-92.

[2] The Seismic Structure of the Sun, Gough DO, Kosovichev AG, Toomre J, Anderson E, Antia HM, Basu S, Chaboyer B, Chitre SM, Christensen-Dalsgaard J, Dziembowski WA, Eff-Darwich A, Elliott JR, Giles PM, Goode PR, Guzik JA, Harvey JW, Hill F, Leibacher JW, Monteiro MJPFG, Richard O, Sekii T, Shibahashi H, Takata M, Thompson MJ, Vauclair S, Vorontsov SV. Science. 1996 May 31;272(5266):1296-300.

[3] The Solar Acoustic Spectrum and Eigenmode Parameters, Hill F, Stark PB, Stebbins RT, Anderson ER, Antia HM, Brown TM, Duvall TL Jr, Haber DA, Harvey JW, Hathaway DH, Howe R, Hubbard RP, Jones HP, Kennedy JR, Korzennik SG, Kosovichev AG, Leibacher JW, Libbrecht KG, Pintar JA, Rhodes EJ Jr, Schou J, Thompson MJ, Tomczyk S, Toner CG, Toussaint R, Williams WE. Science. 1996 May 31;272(5266):1292-6.

[4] Behavior of Li abundances in solar-analog stars. II. Evidence of the connection with rotation and stellar activity. Takeda Y., Honda S., Kawanomoto S., Ando H. And Sakurai T. Astron. Astrophys 2010-1, 515, A93-93.

[5] Xiong Da-run, Deng Li-cai, A re-examination of the dispersion of lithium abundance of Pleiades member stars, Chinese Astronomy and Astrophysics, Volume 30, Issue 1, January-March 2006, Pages 24-40, ISSN 0275-1062.

[6] Eiben, E.; Smith, J. E.: 2003. Introduction to Evolutionary Computation, Springer

[7] Michalewicz, Zbigniew: Genetic Algorithms + Data Structures = Evolution Programs, Springer, 1996.

[8] Javier Causa, Gorazd Karer, Alfredo Nunez, Doris Saez, Igor Skrjanc, Borut Zupancic, Hybrid fuzzy predictive control based on genetic algorithms for the temperature control of a batch reactor, Computers & Chemical Engineering, Volume 32, Issue 12, 22 December 2008, Pages 3254-3263, ISSN 0098-1354

[9] Alfredo Nunez, Doris Saez, Simon Oblak, Igor Skrjanc, Fuzzy-model-based hybrid predictive control, ISA Transactions, Volume 48, Issue 1, January 2009, Pages 24-31, ISSN 0019-0578.

[10] IAU colloquium 121: Inside the sun : Versailles, France, 22-26 May 1989, COSPAR Information Bulletin, Volume 1990, Issue 119, December 1990, Pages 4-5, ISSN 0045-8732

[11] Brian Warner, The abundance of lithium in cool stars, Journal of Quantitative Spectroscopy and Radiative Transfer, Volume 5, Issue 5, September-October 1965, Pages 639-646, ISSN 0022-4073

[12] P. A. Fox, J. M. Fontenla, O. R. White, Solar irradiance variability - comparison of models and observations, Advances in Space Research, Volume 34, Issue 2, Solar Variability and Climate Change, 2004, Pages 231-236, ISSN 0273-1177

[13] Arnab Rai Choudhuri, How far are we from a `Standard Model' of the solar dynamo?, Advances in Space Research, Volume 41, Issue 6, 2008, Pages 868-873, ISSN 0273-1177

[14] Phillip C. Chamberlin, Thomas N. Woods, Francis G. Eparvier, New flare model using recent measurements of the solar ultraviolet irradiance, Advances in Space Research, Volume 42, Issue 5, 1 September 2008, Pages 912-916, ISSN 0273-1177

[15] S. K. Solanki, M. Fligge, Solar irradiance variations and climate, Journal of Atmospheric and Solar-Terrestrial Physics, Volume 64, Issues 5-6, March-April 2002, Pages 677-685, ISSN 1364-6826

[16] A.S. Elasfouri, M.M. Hawas, A simplified model for simulating solar thermal systems, Energy Conversion and Management, Volume 27, Issue 1, 1987, Pages 1-10, ISSN 0196-8904

[17] Alexander Ruzmaikin, John K. Lawrence, Ana Cristina Cadavid, A simple model of solar variability influence on climate, Advances in Space Research, Volume 34, Issue 2, Solar Variability and Climate Change, 2004, Pages 349-354, ISSN 0273-1177

[18] Eugene Eberbach, Toward a theory of evolutionary computation, Biosystems, Volume 82, Issue 1, October 2005, Pages 1-19, ISSN 0303-2647

[19] B. C. Goodwin, Development and evolution, Journal of Theoretical Biology, Volume 97, Issue 1, 7 July 1982, Pages 43-55, ISSN 0022-5193

[20] Zbigniew Michalewicz, Marc Schoenauer, Evolutionary Algorithms, In: Hossein Bidgoli, Editor(s)-in-Chief, Encyclopedia of Information Systems, Elsevier, New York, 2003, Pages 259-267, ISBN 978-0-12-227240-0

[21] Fogel, L. J., Owens, A. J., & Walsh, M. J. 1966, Artificial Intelligence Through Simulated Evolution. New York: Wiley Publishing.

[22] Rechenberg, I. Cybernetic solution path of an experimental problem. 1965, Roy. Aircr. Establ., libr.transl. 1122. Hants, U.K Farnborough.

[23] Rechenberg, I. Evolutionsstrategie: Optimieiung technischer Systeme nach Prinzipien der biologische Evolution. 1973, Stuttgart Frommann-Holzboog.

[24] Schwefel, H.-P. Numerische Optimierung uon Computer-Modellen mittels der Ezdutionsstrategie, 1977, Volume 26 of Interdisciplinary systems research. Basel Birkhauser.

[25] Holland J. Outline for a logical theory of adaptative systems. Journal of the Association for Computing Machinery, (9):297–314, 1962.

[26] John H. Holland. Adaptation in Natural and Artificial Systems. 1975, University of Michigan Press.

[27] Jacco Vink, Supernova remnants with magnetars: Clues to magnetar formation, Advances in Space Research, Volume 41, Issue 3, 2008, Pages 503-511, ISSN 0273-1177

[28] Gibor Basri, William J. Borucki, David Koch, The Kepler Mission: A wide-field transit search for terrestrial planets, New Astronomy Reviews, Volume 49, Issues 7-9, Wide-Field Imaging from Space, November 2005, Pages 478-485, ISSN 1387-6473

[29] F. Selsis, B. Chazelas, P. Borde, M. Ollivier, F. Brachet, M. Decaudin, F. Bouchy, D. Ehrenreich, J.-M. Griessmeier, H. Lammer, C. Sotin, O. Grasset, C. Moutou, P. Barge, M. Deleuil, D. Mawet, D. Despois, J.F. Kasting, A. Leger, Could we identify hot ocean-planets with CoRoT, Kepler and Doppler velocimetry?, Icarus, Volume 191, Issue 2, 15 November 2007, Pages 453-468, ISSN 0019-1035

[30] F. Hechler, W. M. Folkner, Mission analysis for the Laser Interferometer Space Antenna (LISA) mission, Advances in Space Research, Volume 32, Issue 7, Fundamental Physics in Space, October 2003, Pages 1277-1282, ISSN 0273-1177

[31] M. Mayor, D.Queloz, A Jupiter-mass companion to a solar-type star, 1995, Nature, Volume 378, Issue 6555, Pages 355-359.

[32] A. Rozenkiewicz, K. Gozdziewski, Modeling the radial velocities of HD 240210 with genetic algorithms, Proceedings of the 17$^{th}$ young scientists'conference on astronomy and space physics, April 2010

[33] A. Nelder, R.Mead, A simplex method for function minimization, Computer Journal 7, 1965, Pages 308-313.

[34] P. Charbonneau, Genetic algorithms in astronomy and astrophysics, Astrophysical journal supplement series, American Astronomical society, 101, December 1995, Pages 309-334.

[35] T. J. Lazio, J. Cordes, Genetic algorithms on searching for planets around pulsars, ASP Conference Series, Volume 72, 1995, Astronomical Society of Pacific, ISBN 937707-91-0.

[36] A.M. Chwatal, G.R. Raidl, Determining orbital elements of extrasolar planets by evolution strategies, University of Technology, Viena.

[37] R. J. Terrile, C. Adami, H. Aghazarian, S. N. Chau, V.T. Dang, M.I. Ferguson, W. Fink, T.L. Huntsberger, G. Klimeck, M.A. Kordon, S.Lee, P. Allmen, J. Xu, Evolvable Computation Group, Center for Integrated Space Microsystems, Jet Propulsion Laboratory, 4800 Oak Grove Drive Pasadena CA 91109, USA

[38] T. Back, Evolutionary Algorithms in Theory and Practice. Oxford University Press, 1996, New York.

[39] E. Schoneburg, S. Feddersen, Genetische Algorithmen und Evolution-sstrategien. Addison-Wesley, 1994, Berlin.

[40] G.W. Marcy, R.P. Butler, D.A. Fischer, G. Laughlin, S. Vogt, Evidence for Multiple Companions to υAndromedae. Astrophysical Journal 526, 1999, Pages 916–927.

[41] G.W. Marcy, R.P. Butler, D.A. Fischer, G. Laughlin, S. Vogt, A Planet at 5 AU around 55 Cancri. Astrophysical Journal 58, 2002, Pages 1375–1388 Bertiau, F.C., 1957. APJ 125, 696. Srinivas, M., Patnaik, L.M., 1996. IEEE Trans. Knowledge Data Eng. 8, 656.

[42] Charbonneau, P., 2002. An Introduction to genetic algorithms for numerical optimization. http://www.cs.uga.edu/~potter/CompIntell/no_tutorial.pdf. online at May 2011.

[43] Paolo Tanga, D. Hestroffer, M. Delbo, J. Frouard, S. Mouret, W. Thuillot, Gaia, an unprecedented observatory for Solar System dynamics, Planetary and Space Science, Volume 56, Issue 14, Mutual Events of the Uranian Satellites in 2007 - 2008, Mutual Events of the Uranian Satellites in 2007-2008 and Further Observations in Network, November 2008, Pages 1812-1818, ISSN 0032-0633

[44] R. Teixeira, A. Krone-Martins, A. Milone, C. Mallamaci, C. Lopez, J.H. Calderon, I.H. Bustos Fierro, M. Fidencio, R. Zalles, J.L. Muinos, G. Hernandes, T.E.P. Idiart, J.E. Horvath, Contribution to the ground-based follow-up of the Gaia space mission, Planetary and Space Science, Volume 56, Issue 14, Mutual Events of the Uranian Satellites in 2007 - 2008, Mutual Events of the Uranian Satellites in 2007-2008 and Further Observations in Network, November 2008, Pages 1828-1831, ISSN 0032-0633

[45] M.D. Smith, The origin of stars, London Imperial College Press, 2004, ISBN 1860945015

[46] A. G. Kosovichev, Probing solar and stellar interior dynamics and dynamo, Advances in Space Research, Volume 41, Issue 6, 2008, Pages 830-837, ISSN 0273-1177

[47] P. Bodenheimer, Stellar Structure and Evolution, In: Robert A. Meyers, Editor(s)-in-Chief, Encyclopedia of Physical Science and Technology, Academic Press, New York, 2001, Pages 45-78, ISBN 978-0-12-227410-7

[48] R. Wilson, E. Devinney, 1971, ApJ, 166, 605.

[49] C. Stagg, E. Milone, 1993, in Light Curve Modeling of Eclipsing Binary Stars, ed. E.F. Milone, New York: Springer-Verlag, 75.

[50] T.S. Metcalfe, Genetic-algorithm-based light-curve optimization applied to observations of the W Ursae Majoris star BH Cassiopeiae, The Astronomical Journal, 1999

[51] H.L.Johnson, W. Morgan, 1953, Fundamental stellar photometry for standards of spectral type on the revised system of the Yerkes spectral atlas, The Astrophysical Journal, vol. 117, pp. 313-352.

[52] I. Iben, "Planetary nebulae and their central stars - origin and evolution", Physics Reports 250, 1995, l-94.

[53] S. Turck-Chièze, w. Däppen, E.Fossat, J.Provost, E.Schatzman, D.Vignaud, "The solar interior", PHYSICS REPORTS (Review Section of Physics Letters) 230, Nos.2,4, 1993, 52,235.

[54] E. Alba, Parallel evolutionary algorithms can achieve super-linear performance, Information Processing Letters, (82):7–13, 2002.

[55] S.Kawaler, V.Trimble, Stellar Interiors - Physical Principles, Structure, and Evolution, Springer, 2004.

[56] S.Kawaler, W.Bohn-Vitense Introduction to Stellar Astrophysics, Volume 2: Stellar Atmospheres, Cambridge, 2004, Cambridge University Press.

[57] S.D. Kawaler, A.Bradley, Precision asteroseismology of the pre-white dwarf star PG 1159-035, 1994, Astrophysical Journal.

[58] Empirical modeling of the solar corona using genetic algorithms. SE Gibson, P Charbonneau Journal of Geophysical Research 103, 14, 7/1998

[59] Tomczyk, S., Streander, K., Card, G., Elmore, D., Hull, H. & Cacciani, A. 1995b, Solar Physics.

[60] Lazio T., Genetic Algorithms, Pulsar Planets, and Ionized Interstellar Microturbulence, Thesis (PHD). Cornell University, Source DAI-B 58/04, p. 1925, Oct 1997, 414 pages.

[61] Goździewski K., Maciejewski A. J., Migaszewski C., 2007, On the Extrasolar Multiplanet System around HD 160691, The Astrophysical Journal 657: 546-558.

[62] Wahde M.,Donner K.J. 2001, Determination of the orbital parameters of the M 51 system using a genetic algorithm, A&A, 379, 115–124

[63] Theis C., Spinneker C., M51 revisited, a genetic algorithm approach to its interaction history, Kluwer academic publishers.

[64] Cantú-Paz E: Evolving neural networks for the classification of galaxies. In Proceedings of the Genetic and Evolutionary Algorithm Conference. Edited by Langdon, WB, Cantu-Paz E, Mathias K, Roy R, Davis D, Poli R, Balakrishnan K, Honavar V, Rudolph G, Wegener J, Bull L, Potter MA, Schultz AC, Miller JF, Burke E, Jonoska. San Francisco, Morgan Kaufman Publishers; 2002:1019-1026

[65] Kippenhahn R., Weigert A., Stellar Structure and Evolution, Berlin,1994, Springer

[66] R.M. Lark, Spatially nested sampling schemes for spatial variance components: Scope for their optimization, Computers & Geosciences, In Press, Corrected Proof, Available online 10 March 2011, ISSN 0098-3004

[67] Salvador Garcia, Joaquin Derrac, Julian Luengo, Cristobal J. Carmona, Francisco Herrera, Evolutionary selection of hyperrectangles in nested generalized exemplar learning, Applied Soft Computing, Volume 11, Issue 3, April 2011, Pages 3032-3045, ISSN 1568-4946

[68] Jeongwen Chiang, Siddhartha Chib, Chakravarthi Narasimhan, Markov chain Monte Carlo and models of consideration set and parameter heterogeneity, Journal of Econometrics, Volume 89, Issues 1-2, 26 November 1998, Pages 223-248, ISSN 0304-4076

[69] D.Ordóñez, C. Dafonte, M. Manteiga, Parameterization of RVS synthetic stellar spectra for the ESA Gaia mission: Study of the optimal domain for ANN training. Expert Syst. Appl. 37(2),2010, 1719-1727

[70] C. Theis, Modeling Encounters of Galaxies: The Case of NGC4449, Proc. of the Annual Meeting of the Astronomische Gesellschaft, Heidelberg, 1998, Rev. Mod. Astron. 12

[71] M. Wahde, K.J. Donner, Determination of the orbital parameters of the M 51 system using a genetic algorithm, 2001, A&A 379, 115-124

[72] P. Zwart, T. Totani, Precessing jets interacting with interstellar material as the origin for the light curves of gamma-ray bursts, Monthly Notices of the Royal Astronomical Society, Volume 320, pags 951-957, December 2001

[73] M. Lightman, J. Thurakal, J.Dwyer, Prospects of gravitational wave data mining and exploration via evolutionary computing, 2006, Physics Rev., Lertt 97,151101.

[74] J. Crowder, N.L. Cornish, J.L. Reddinger, LISA data analysis using genetic algorithms, 2006, Physics Rev., D73, 063011.

[75] J.R. Gair, E.K. Porter, Cosmic Swarms: A search for Supermassive Black Holes in the LISA data stream with a Hybrid Evolutionary Algorithm, 2010, Physics Rev., D81, 104014.

[76] A. Petiteau, Y. Shang, S. Babak, The search for black holes binaries using a genetic algorithm, 2009, Classical and quantum gravity, 26, 204011

[77] F. Attia, E. Mahmoud, H.I. Shahin, A modifies genetic algorithm for precise determination the geometrical orbital elements of binary stars, 2009, New astronomy, Volume 14, Issue 3, pags 285-293

[78] T.S. Metcalfe, P. Charbonneau, Stellar structure modeling using a parallel genetic algorithm for objective global optimization, 2003, Journal of Computational Physics, 185, 176-193

[79] Dorigo, M. y Gambardella, L.M A cooperative learning approach to the traveling salesman problem, 1997, IEEE Transactions on Evolutionary Computation, 1, 53-66.

[80] Dorigo, M. y Stützle, T. Ant Colony Optimization, 2004, Cambridge: MIT Press.

[81] Jennings, N.R On agent-based software engineering, 2000, Artificial Intelligence, 117, 277-296.

[82] Zadeh, L. A. Fuzzy logic, neural networks, and soft computing, 1994, Communications of the ACM, 37, 77-84.

[83] Tagliaferri R. Introduction: Neural networks for analysis of complex scientific data: Astronomy and geosciences, Neural Networks, 2003, Volume 16, 3-4.

[84] Tagliaferri R.Neural Nerwoks in Astronomy, Neural Networks, 2003, Volume 16, 3-4.

[85] Gibson, S.E., Charbonneau, P. Empirical modeling of solar corona using genetic algorithms, Space Physics, Volume 103, Issue A7, Pags 14511-14521.

[86] Charbonneau, P. The solar corona as a minimum energy system, Solar Physics, Volume 165, Issue 2, Pags 237-256.

[87] Koza J.R, Genetic Programming: A paradigm for genetically breeding populations of computer programs to solve problems.

[88] Lewis, M.A., Genetic Programming approach to the construction of a neural network for control of a walking robot, Robotics and Automation, 1992, Volume 3, pags 2618-2623.

[89] Grant, E., Incremental evolution of autonomous controllers for unmanned aerial vehicles using multi-objective genetic programming, Cibernetics and intelligent System, 2004, Volume 2, pags 689-694.

[90] Smith, J.F., Autonomous and cooperative robotic behavior based on fuzzy logic and genetic programming, Integrated Computer-Aided Engineering, Volume 14, pags 141-159.

[91] Floreano, D., Evolutionary neurocontrollers for autonomous mobile robots, Neural networks, 1998, Volume 11, Pags 1461-1478

[92] Clayton, D.D, Principles of stellar evolution and nucleosynthesis, 1968, Harvard University Press

[93] Kippenhahn, R. Stellar structure and evolution, 1994, Springer-Verlag Berlin Heidelberg New York

[94] Daida,J.M., Evolving Feature-extraction algorithms: Adapting genetic programming for image analysis in geoscience and remote sensing, Proceeings of the 1996 international geoscience and remote sensing symposium, IEEE Press.

[95] Smart, W.R., Classification strategies for image classification in genetic programming, Proceeding of image and vision, 2003.

[96] Chwatal A.M, Fitting Multi-Planet transit models to photometric time-data series by evolution strategies, 12[th] annual conference on Genetic Computation.

[97] Calvo, R.A., H.A. Ceccatto, R.D. Piacentini: Neural Network Prediction of Solar Activity, 1995, The Astrophysical Journal, 444, Pags 916-921

[98] Conway, A.J., K.P. Macpherson, G.Blacklaw, J.C.: A neural network prediction of solar cycle 23, Journal of Geophysical Research, 1988, vol.103, Pags 733-29, 742

[99] Jagielski, R., Genetic Programming Prediction of Solar Activity, Lecture Notes in Computer Science, 2000, Volume 1983, 2000, Pages 191-210

[100] Strangeways, H., Kutiev, I, Near-Earth space plasma modeling and forecasting, Annals of geophysics, Volume 52, pags 255-260.

[101] Teuben, P., The Stellar Dynamics Toolbox NEMO, Astronomical Data Analysis Software and Systems IV, ASP Conference Series, Vol. 77, 1995, pag 398

[102] Teuben, P., Genetic Programming and other fitting techniques in Galactic Dynamics, Data Analysis Software and Systems (ADASS),2004, Volume 314, Pag 621

[103] Li J., Yao X., Frayn C., An Evolutionary Approach to Modeling Radial Brightness Distributions in Elliptical Galaxies, PARALLEL PROBLEM SOLVING FROM NATURE, Lecture Notes in Computer Science, 2004, Volume 3242, Pags 591-601

[104] Bergh, S.V, Galactic and Extragalactic Supernova Rates, Annual Review of Astronomy and Astrophysics, Volume 29, pags 363-407

[105] Mesot J., Janssen S., Focus: Project of a Space and Time Focussing Time-of-Flight Spectrometer for Cold Neutrons at the Spallation Source SINQ of the Paul Scherrer Institute, Journal of Neutron Research, 1996, Volume 3, Pages 293-310

[106] Link J.M., Yager P.M., Nuclear Instruments and Methods in Physics Research Section A: Accelerators, Spectrometers, Detectors and Associated Equipment, Volume 551, Issues 2–3, 11 October 2005, Pages 504–527